# Leveraging WaveNet for Dynamic Listening Head Modeling from Speech

Minh-Duc Nguyen
Department of Artificial Intelligent Convergence
Chonnam National University
Gwangju, Republic of Korea
ducnm@jnu.ac.kr

Hyung-Jeong Yang[*]
Department of Artificial Intelligent Convergence
Chonnam National University
Gwangju, Republic of Korea
hjyang@jnu.ac.kr

Seung-Won Kim
Department of Artificial Intelligent Convergence
Chonnam National University
Gwangju, Republic of Korea
Seungwon.Kim@jnu.ac.kr

Ji-Eun Shin
Department of Psychology
Chonnam National University
Gwangju, Republic of Korea
jieunshin@jnu.ac.kr

Soo-Hyung Kim
Department of Artificial Intelligent Convergence
Chonnam National University
Gwangju, Republic of Korea
shkim@jnu.ac.kr

*Abstract*— The creation of listener facial responses aims to simulate interactive communication feedback from a listener during a face-to-face conversation. Our goal is to generate believable videos of listeners' heads that respond authentically to a single speaker by a sequence-to-sequence model with an combination of WaveNet and Long short-term memory network. Our approach focuses on capturing the subtle nuances of listener feedback, ensuring the preservation of individual listener identity while expressing appropriate attitudes and viewpoints. Experiment results show that our method surpasses the baseline models on ViCo benchmark Dataset.

## I. Introduction

Face-to-face interaction refers to the communication and relationship dynamics between two or more individuals, typically in a face-to-face setting. This form of interaction is fundamental to human social behavior, as it encompasses a wide range of interpersonal activities, from casual conversations to deep emotional exchanges. Understanding dyadic interaction is crucial for fields such as psychology, sociology, communication studies, and even artificial intelligence, where the nuances of two-person interactions can offer insights into human behavior, social bonding, and communication patterns. Additionally, real-time analysis of facial expressions in face-to-face interactions is essential for enhancing emotional comprehension. Investigating this through computer vision, particularly with dynamic talking human videos, presents an intriguing challenge. Creating responsive listener reactions is key to achieving realistic digital human interactions across various applications, ranging from human-computer interaction to animation production.

Most previous works concentrate on speaker modeling, specifically on generating talking faces. These efforts have aimed to create lifelike speaker animations by synchronizing lip movements, facial expressions, and speech. Some studies [10, 11, 12] focus on creating realistic facial animations by directly using input audio to drive the lip movements of the target speaker, leveraging the strong correlation between speech and lip motion. Besides the task of talking face generation, the generation of listener reactions remains largely unexplored. Listener reaction modeling is crucial for realistic digital interactions, as it involves creating responsive facial expressions and head movements that accurately reflect the listener's engagement and emotional responses. Developing such models could significantly enhance applications in human-computer interaction, virtual reality, and animation, providing a more immersive and authentic communication experience. While the strong correlation between speech and lip motion has been leveraged for speaker modeling, understanding and replicating the nuanced feedback from listeners presents a new and intriguing challenge. Mohan Zhou, et al. [1] introduce paired videos of speakers and listeners datasets for listening head generation task. To facilitate the generation of listener head movements, they employed an LSTM-based [3] model to process input signals and output features related to the listener's head pose. Subsequently, PCHG [2] enhanced the generated frames using a face segmentation model to improve background stability. Despite their ability to capture short and moderate-term dependencies in sequences, their approaches with LSTMs can struggle with learning and remembering dependencies that span very long sequences. This can limit their effectiveness in tasks requiring extensive context understanding over extended periods.

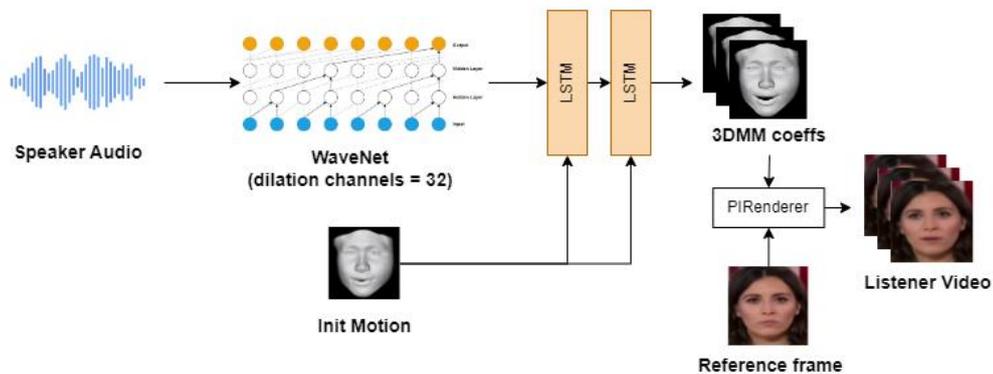

Fig. 1. The pipeline of proposed model.

In this paper, we tackle the challenge of managing long-term dependencies in sequence modeling, we integrate the WaveNet [4] architecture as a foundation for extracting speech representations from surface features. WaveNet excels in learning high-level, speaker-independent representations of speech directly from raw audio waveforms, capturing fine details and nuances in speech patterns. We combine it with LSTM neural networks to predict the listener outcome. LSTM with WaveNet head model can better learn the innate long-term and short-term relationships in sequence data and is more adept at capturing the broad trend. Our objective is to train a model that translates audio into listener reactions, utilizing a powerful renderer to generate vivid videos.

## II. RELATED WORKS

Modeling listening behaviors has been a less explored area compared to speaking in both applications and research literature. The first work in [6] introduced a novel approach utilizing data-driven techniques to create an animated character capable of reacting to the speaker's voice. The listening head generation task [1, 15] focuses on predicting facial expressions in response to non-verbal cues. Recent scholarly investigations into facial reactions [1, 16] frequently utilize 3D Morphable Model (3DMM) [5] coefficients, such as those for expressions and poses, to enhance the fidelity of facial muscle movement visualization. Moreover, these studies integrate the speaker's speech patterns and audio behaviors as supplementary inputs, enriching the portrayal of both verbal and non-verbal aspects of the speaker's behavior.

In this work, we propose a deep neural network model that combines WaveNet and LSTM, incorporating both recurrent and 1D dilated methods to identify longer trends in speaker audio input sequence and predict dynamic listener head motions as 3DMM parameters.

## III. METHOD

Our objective is to generate comprehensive dynamic listening head videos from a given speaker's audio information and a listener head image. The WaveNet-LSTM architecture model first predicts the listener's head motion and facial expression features. These predicted features are adjusted to reconstruct the 3D Morphable Model (3DMM) [5] coefficients, incorporating the past motion features of the reference listener. Subsequently, the adapted coefficients are fed into PIRender [13], a neural renderer, to generate a listening reaction video.

### A. Model Architecture

We implement WaveNet-like architecture as our deep speech feature extractor. WaveNet is a type of autoregressive generative model designed to directly analyze time-series data. It achieves this by modeling the conditional probability distribution of the time-series data through a series of 1D dilated causal convolution layers stacked together. This approach allows WaveNet to predict each data point in the sequence based on previous data points, without needing explicit alignment or segmentation.

For a given time series $x = \{x_1, …, x_T\}$, WaveNet factorizes its joint probability as a product of its conditional probabilities. This means that the model learns to predict each data point $x_t$ based on all previous data points $x_1, …, x_{t-1}$, capturing the sequential dependencies inherent in the data as shown in the following equation:

$$p(x) = \prod_{t=1}^{T} p(x_t | x_1, …, x_{t-1}) \qquad (1)$$

The core component of WaveNet models is the dilated causal convolution. This technique preserves the sequential order of the data, ensuring that the model respects the temporal structure. In a layer with dilated causal convolutions, filters are applied to the input sequence by skipping a fixed number of steps, determined by the dilation rate. The dilation rate grows exponentially with each subsequent layer, allowing the model to achieve exponentially larger receptive fields as it goes deeper as illustrated in Fig. 2. In addition to model complex operations, WaveNet uses gated activations, the residual block and skip connections architecture which are indicated in Fig. 3. The gated activation units is represented by the following equation:

$$z = \tanh(W_f * x) \odot \sigma(W_g * x) \quad (2)$$

where $*$ is a convolution operator, $\odot$ is an element wise multiplication operator, $\sigma(.)$ is the sigmoid activation function, k is the layer index and $W_f$, and $W_g$, are weight matrix of filters and gate respectively.

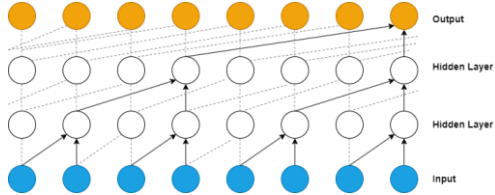

Fig. 2. Dilated causal convolution with dilation rate of 1, 2 and 4.

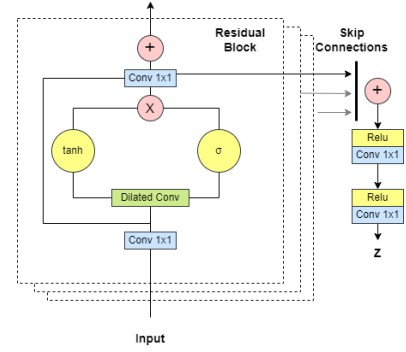

Fig. 3. Residual block with three layers.

In our proposed model illustrated in Fig. 1, we incorporate two LSTM layers at the end of the WaveNet model. This design enables our model to effectively learn both long-term and short-term time series dependencies from the input data. Additionally, a reference image of listener head which considered as listener initial head motion is embedded into each LSTM layer's hidden state.

### B. Loss Function

We compute $L_2$ distance separately within head pose and expression which are encompassed in 3DMM coefficients set. The loss is computed as:

$$\mathcal{L} = \sum_{t=1}^{T} |\beta_t - \hat{\beta}_t|_2 + |c_t - \hat{c}_t|_2 + |p_t - \hat{p}_t|_2 + |\mu(c_t) - \mu(\hat{c}_t)|_2 \quad (3)$$

where $\beta_t$, $c_t$, $p_t$ and $\hat{\beta}_t$, $\hat{c}_t$, $\hat{p}_t$ denote the ground truth and predictions of angle, translation, and expression coefficients respectively. The head motion inter-frame changes denote as $\mu(.)$.

## IV. EXPERIMENT

### A. Dataset

We evaluate our approach using The ViCo dataset [1], which comprises 483 video clips depicting real-world face-to-face interactions involving 67 speakers and 76 listeners across various natural settings. The dataset includes diverse samples categorized into three main attitude categories: Positive, Natural, and Negative. Following the methodology outlined in PIRender [13], we extract 3D Morphable Model (3DMM) coefficients—encompassing identity, expression, texture, pose, and lighting—from the videos at a frame rate of 30 frames per second, with each face video frame resized to 256 x 256 pixels. Additionally, 45-dimensional acoustic features containing: 14-dim Mel-Frequency Cepstral Coefficients (MFCC), 28-dim MFCC-Delta, Zero Crossing Rate (ZCR), loudness, and energy are extracted from the audio data.

### B. Evaluation Metrics

Our evaluation aligns with ViCo [1] benchmark methodology. To assess the accuracy of generated pose and expression features, we employ the *L1* distance as our chosen evaluation metric. For a comprehensive evaluation of video-level performance, we utilize a diverse set of metrics, including Structural Similarity (SSIM) [14], Cumulative Probability of Blur Detection (CPBD) [9], Peak Signal-to-Noise Ratio (PSNR), and Fréchet Inception Distance (FID) [7]. To evaluate identity preservation, we measure the cosine similarity (CSIM) between the identity of generated images and ground truth images.

### C. Results

We compare our methods with two existing approaches for responsive listener head generation: the LSTM-based sequential decoder from ViCo and PCHG [2], which improves created films by stabilizing the background with a segmentation model. We conducted the training of both models on ViCo training set $\mathcal{D}_{train}$ and were evaluated on the test set $\mathcal{D}_{test}$ and an out-of-domain set $\mathcal{D}_{ood}$. Specifically, all identities in $\mathcal{D}_{test}$ have appeared in $\mathcal{D}_{train}$, while identities in $\mathcal{D}_{ood}$ have no overlap with those in $\mathcal{D}_{train}$. In terms of feature distance, as shown in Table I, we further compare with the following baselines provided by [1]: Random generates frames from a reference image with small perturbations to mimic random head motion; Simulation simulates natural listening behavior by repeating motion patterns from $\mathcal{D}_{train}$; Simulation* repeats natural listening motion patterns from $\mathcal{D}_{train}$ with the corresponding attitude. Our method achieves the best results on all there feature distances of angle/expression/translation coefficients respectively. Additionally, Table II demonstrates the evaluation results of zero-shot 3D face rendering [13], our model gets the best overall performance on those video-level criteria. The results in Fig. 4 are derived from the $\mathcal{D}_{test}$ and $\mathcal{D}_{ood}$ as our qualitative visualization. Our method successfully generates dynamic reaction video clips of listening heads with a reasonable level of quality.

TABLE I. FEATURE DISTANCE RESULTS OF CE (X100) EVALUATION

| Method | Test set | Average of all Attitudes motion | | |
|---|---|---|---|---|
| | | angle ↓ | exp ↓ | trans ↓ |
| Random | $D_{test}$ | 18.04 | 44.67 | 19.80 |
| | $D_{ood}$ | 18.11 | 44.60 | 20.36 |
| Simulation | $D_{test}$ | 10.81 | 7.37 | 13.52 |
| | $D_{ood}$ | 9.91 | 28.66 | 11.76 |
| Simulation* | $D_{test}$ | 11.24 | 29.20 | 11.00 |
| | $D_{ood}$ | 12.58 | 28.46 | 11.55 |
| ViCO | $D_{test}$ | 7.79 | 15.04 | 6.52 |
| | $D_{ood}$ | 8.23 | 22.83 | 8.32 |
| PCHG | $D_{test}$ | 13.37 | 19.98 | 7.74 |
| | $D_{ood}$ | 18.00 | 18.82 | 8.76 |
| Ours | $D_{test}$ | **8.17** | **13.39** | **5.27** |
| | $D_{ood}$ | **4.90** | **17.62** | **6.32** |

TABLE II. QUANTITATIVE EVALUATIONS ON RENDERED FRAMES

| Method | SSIM ↑ | CPBD ↑ | PSNR ↑ | FID ↓ | CSIM ↓ |
|---|---|---|---|---|---|
| ViCO [1] | 0.56 | 0.11 | 17.36 | 27.74 | 0.23 |
| PCHG [2] | 0.58 | 0.16 | 18.51 | **21.35** | 0.25 |
| Ours | **0.62** | **0.17** | **18.68** | 27.11 | **0.06** |

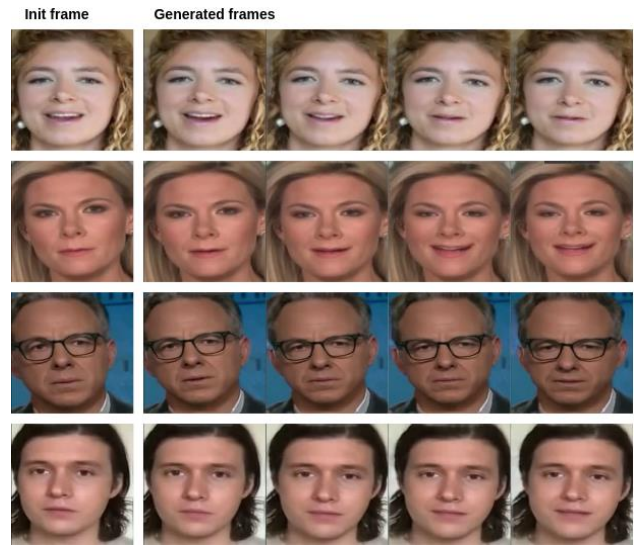

Fig. 4. Visualization of generated listening heads.

## V. CONCLUSION

In this paper, we present a deep neural network model to predict the head motion of the listener using speaker acoustic audio from the ViCo dataset. We demonstrate that the proposed WaveNet + LSTM model performs best compared to other previous methods. The model effectively captures both short-term and long-term dependencies in speaker audio, enabling it to discern patterns and generate dynamic coefficients for listening head movements. The approach underwent comprehensive evaluations, and both quantitative and experimental results confirmed its superior ability to generate precise listener motion responses.

## *Acknowledgment*

This work was supported by the National Research Foundation of Korea (NRF) grant funded by the Korea government (MSIT) (RS- 2023- 00219107), in part by the Institute of Information and Communications Technology Planning and Evaluation (IITP) through the Artificial Intelligence Convergence Innovation Human Resources Development grant funded by the Korea Government (MSIT) under Grant IITP-2023-RS-2023-00256629, and in part by the MSIT (Ministry of Science and ICT), Korea, under the Innovative Human Resource Development for Local Intellectualization support program (IITP-2023-RS-2022-00156287) supervised by the IITP(Institute for Information & communications Technology Planning & Evaluation).